\documentclass[10pt,twocolumn,letterpaper]{article}

\usepackage{iccv}
\usepackage{times}
\usepackage{epsfig}
\usepackage{graphicx}
\usepackage{amsmath}
\usepackage{amssymb}
\usepackage{multirow}
\usepackage{float}
\usepackage{placeins}

\usepackage[pagebackref=true,breaklinks=true,letterpaper=true,colorlinks,bookmarks=false]{hyperref}

\iccvfinalcopy 

\def\iccvPaperID{****} 

\ificcvfinal\fi
\begin{document}

\title{Unsupervised learning from video \\ to detect foreground objects in single images}

\author{\begin{tabular}{ccc}
Ioana Croitoru\textsuperscript{1} & Simion-Vlad Bogolin\textsuperscript{1} & Marius Leordeanu\textsuperscript{1,2}\\
{\tt\small ioana.croi@gmail.com} & {\tt\small vladbogolin@gmail.com} & {\tt\small marius.leordeanu@imar.ro}\\
\end{tabular}\\
\begin{tabular}{cc}
\textsuperscript{1}Institute of Mathematics of the Romanian Academy & \textsuperscript{2}University "Politehnica" of Bucharest \\
{\small 21 Calea Grivitei, Bucharest, Romania} & {\small 313 Splaiul Independentei, Bucharest, Romania}\\
\end{tabular}
}

\maketitle

\begin{abstract}
    Unsupervised learning from visual data is one of the most difficult challenges in computer vision, 
    being a fundamental task for understanding how visual recognition works. 
    From a practical point of view, learning from unsupervised visual input has an immense practical value, as very large quantities of unlabeled videos can be collected at low cost. In this paper, we address the task of unsupervised learning to detect and segment foreground objects in single images. We achieve our goal by training a student pathway, consisting of a deep neural network. It learns to predict from a single input image (a video frame) the output for that particular frame, of a teacher pathway that performs unsupervised object discovery in video. Our approach is different from the published literature that performs unsupervised discovery in videos or in collections of images at test time. We move the unsupervised discovery phase during the training stage, while at test time we apply the standard feed-forward processing along the student pathway. This has a dual benefit: firstly, it allows in principle unlimited possibilities of learning and generalization during training, while remaining very fast at testing. Secondly, the student not only becomes able to detect in single images significantly better than its unsupervised video discovery teacher, but it also achieves state of the art results on two important current benchmarks, YouTube Objects and Object Discovery datasets. Moreover, at test time, our system is at least two orders of magnitude faster than other previous methods. 
\end{abstract}

\section{Introduction}
  
   The problem of unsupervised learning is one of the most difficult and intriguing in computer vision and machine learning today. In a very general sense, many researchers believe that unsupervised learning from video could help decode many hard questions regarding the nature of intelligence and learning. Since unlabeled videos are easy to collect at a very low cost, solving this task would bring a great practical value in many vision and robotics applications. There are several papers addressing this difficult task, but the current methods are still far from fully solving the challenge. Many recent unsupervised methods in vision follow two main directions: one is to learn powerful features in a completely unsupervised manner and then use them in a classic supervised learning scheme in combination with different classifiers, such as SVMs or CNNs~\cite{radenovic2016cnn, misra2016shuffle, li2016unsupervised}. The second, more classical line of research, is to discover common patterns in unlabeled data, at test time, using different clustering, feature matching or other data mining approaches~\cite{jain1999data,cho2015unsupervised,key:sivic_05}. In the first case the unsupervised learning task is limited to the intermediate level of feature learning, while in the second, its performance depends on the specific structure of the image collection given at test time. 
   
   The task of object discovery and unsupervised learning in video is related to co-segmentation~\cite{joulin2010discriminative,kim2011distributed,rubinstein2013unsupervised,joulin2012multi,kuettel2012segmentation,vicente2011object,rubio2013video} and weakly supervised localization~\cite{deselaers2012weakly,nguyen2009weakly,siva2013looking}. Earlier methods are based on local feature matching and detection of their co-occurrences patterns~\cite{stretcu2015multiple,key:sivic_05,key:leordeanu_cvpr05,key:parikh_07_2,liu2007topic}, while recent approaches~\cite{joulin2014efficient,rochan2014efficient} discover object tubes by linking candidate detections between frames with or without refining their location. Traditionally, the task of unsupervised learning from image sequences, formulated as an optimization problem for either feature matching, conditional random fields or data clustering is inherently expensive due to the combinatorial nature of the problem. That is why our approach, in which we learn to detect in a fast, feed-forward manner from an unsupervised object discoverer in video (while having virtually unlimited training data), has certain advantages that might open new possibilities in the quest for solving the unsupervised learning problem in the real world. 
   
\begin{figure}[t]
\begin{center}
   \includegraphics[width=1\linewidth]{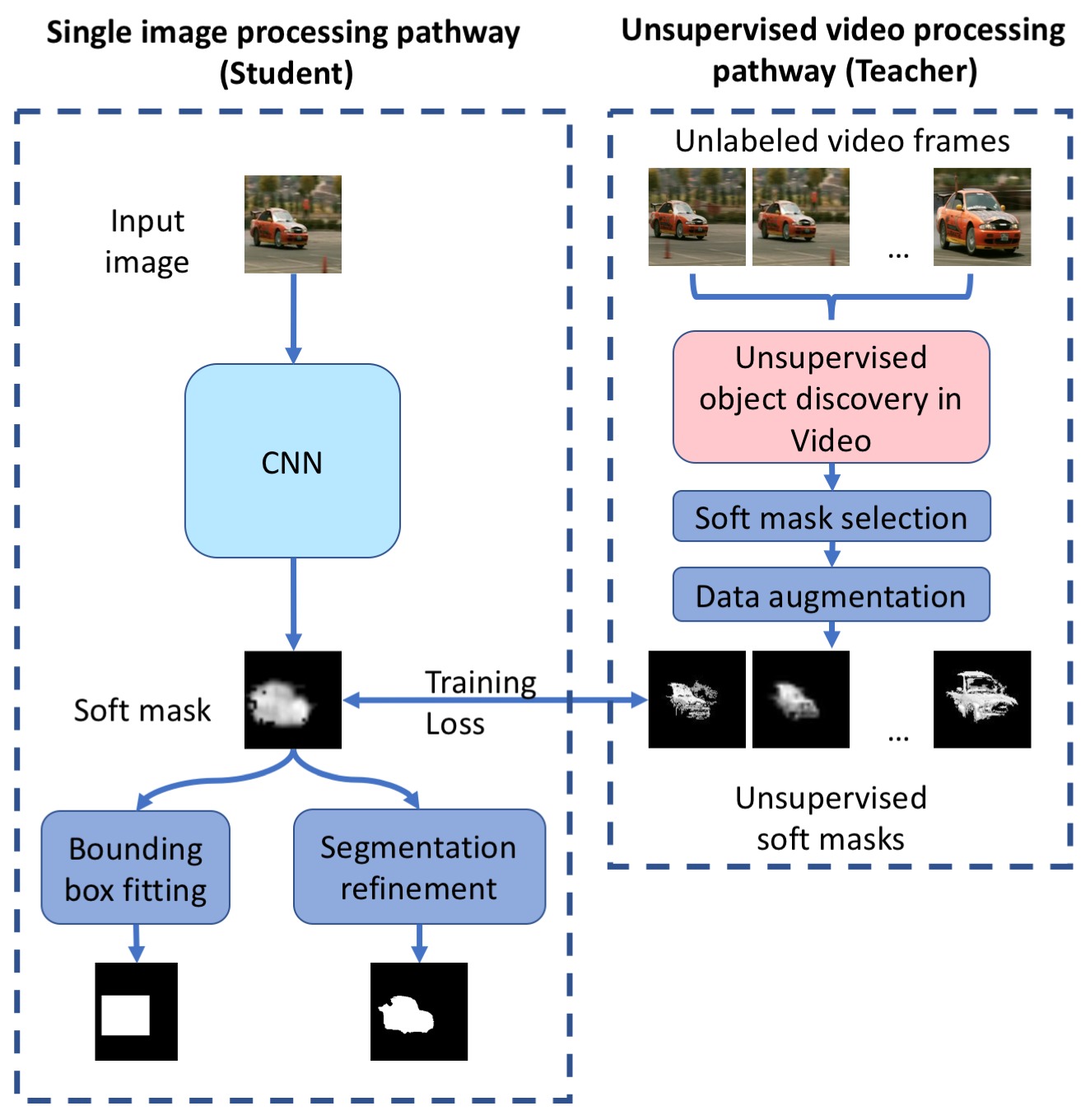}
\end{center}
   \caption{Our dual student-teacher system used for unsupervised learning to detect foreground objects in images. It combines two processing pathways: the teacher, on the right hand side, discovers in an unsupervised fashion foreground objects in video and outputs soft masks for each frame. The resulting soft masks, are then filtered and only good segmentations are kept, based on a simple and effective unsupervised quality metric.
   The set of selected segmentations is then augmented in a relatively simple manner, automatically. The resulting final set of pairs - input image (a video frame) and soft mask (the mask for that particular frame which acts as an unsupervised label) - are used for training the student CNN pathway. Note that the student, after being fully trained, outperforms the teacher.
   }
\label{fig:system}
\end{figure}

   Our system is presented in Figure \ref{fig:system}.
   We have an unsupervised training stage, in which a student deep neural network (Figure \ref{fig:network}) learns frame by frame from an unsupervised teacher, which performs object segmentation discovery in videos, to produce similar object masks in single images. The teacher method takes advantage of the consistency in appearance, shape and motion manifested by objects in video. In this way, it discovers objects in the video and produces a foreground segmentation for each individual frame. Then, the student network tries to imitate for each frame the output of the teacher, while having as input only a single image - the current frame. The teacher pathway is much simpler in structure, but it has access to information over time. In contrast, the student is much deeper in structure, but has access only to one image. Thus, the information discovered by the teacher in time is captured by the student in depth, over neural layers of abstraction. In experiments, we show a very encouraging fact: the student easily learns to outperform its teacher and discovers by itself general knowledge about the shape and appearance properties of objects, well beyond the abilities of the teacher. Thus, the student produces significantly better object masks, which generally have a good form, do not have holes and display smooth contours, while having an appearance that is often in contrast to the background scene. 
   
  Since there are available methods for video discovery with good performance, the training task becomes immediately feasible. In this work we chose the VideoPCA algorithm introduced as part of the system in~\cite{stretcu2015multiple} because it is very fast (50-100 fps), uses very simple features (pixel colors) and it is completely unsupervised - with no usage of supervised pre-trained features. That method exploits the stability in appearance and location of objects, which is common in video shots.  
  While the object masks discovered are far from being perfect and are often noisy, the student deep network manages to generalize and overcome some of these limitations. We propose a ten layer deep neural network for the student pathway (Figure \ref{fig:network}). It takes as input the original RGB, HSV and image spatial derivatives channels. It outputs a low resolution soft segmentation mask of the main objects present in a given image.

\paragraph{Main contributions:}
Our main contributions are: 

\textbf{1)} Our approach, to our best knowledge, is the first one that learns to detect and segment foreground objects in images in a completely unsupervised fashion, with no pre-trained features needed or manual labeling, while requiring only a single image at test time.

\textbf{2)} We propose a novel architecture for unsupervised learning in video, consisting of two processing pathways, with complementary functions and properties. The first pathway discovers foreground objects in videos in an unsupervised manner and has access to all the video frames. It acts as a teacher. The second "student" pathway, which is a deep convolutional net, learns to predict the teacher's output for each frame while having access only to a single input image. An important fact, shown in our experiments, is that the student learns to outperform its teacher, despite being limited to a single image input. Once trained using our dual-pathway system, the student achieves state of the art results on two important datasets.

\section{Our approach and intuition}

There are several observations that motivate the approach we take for addressing the unsupervised learning task. First, we notice that unsupervised learning methods are generally more effective when considering video input, in which objects satisfy spatio-temporal consistency, with smooth variations in shape, appearance and location over time. For that matter it is usually harder to learn about objects from collections of images that are independently taken. 
This motivates the inclusion of the video discovery pathway, for which there are available several published methods that could be used. We should also keep in mind that the video discovery module should not use pre-trained features on manually labeled ground truth, if we want to develop a fully unsupervised method. On the contrary, the visual cues and features used by the teacher should be as simple as possible, such as individual pixel colors. These are precisely the features used by the teacher pathway of our choice, the VideoPCA algorithm introduced in~\cite{stretcu2015multiple}. That method has the added quality of being very fast and reasonably accurate.

Second, if we want the student pathway to learn general principles about objects in images, we need to limit its access to a single input image. Otherwise, if given the entire video as input, a powerful deep network would easily overfit when trained to predict the teacher's output.

The most important question that needs to be answered is whether the student can outperform its teacher. If this is indeed the case, then the student, processing a single image, has an important quality, besides the speed advantage over the teacher (which needs to process an entire video). Once the student progresses beyond the capabilities of its teacher, we could indeed envision the potential practical advantages of unsupervised learning - especially when there is so much unlabeled video data available. Therefore, we first have to make sure that the student receives only the best quality input possible from the teacher. For that we add an extra module for unsupervised soft masks selection. It is based on a simple and intuitive measure of quality, explained in detail later, which does a good job at ordering masks with respect to their true quality. 

Then, we also need to make sure that the student sees as much training data as possible. For that, we design an automatic data augmentation module, which creates extra training data by randomly scaling and shifting the masks provided by the teacher after the mask selection procedure.

Having this in mind, one of the important findings in our experiments is that in all our tests
the student indeed outperforms its teacher. Moreover, it achieves state of the art results on two important and different benchmarks. We believe that the success of this unsupervised learning paradigm is due to the fact that the student is forced to capture from appearance only (as it is limited to a single image) visual properties and cues that are good predictors for the presence of objects.

\section{System architecture}

We now present in more detail the architecture of our system, module by module, as it is presented in Figure \ref{fig:system}.

\subsection{Teacher path: unsupervised discovery in video}

There are several methods already
available for discovering objects and salient regions in images and videos~\cite{borji2012salient,cheng2015global,hou2007saliency, jiang2013salient,cucchiara2003detecting,barnich2011vibe},
with reasonably good performance. More recent methods for foreground objects discovery such as~\cite{papazoglou2013fast,stretcu2015multiple} are both relatively fast and accurate, with testing time above $4$ seconds per frame. However, this time is still long and prohibitive for training a deep neural net (the student pathway) that requires millions of images to train. For that reason we chose the VideoPCA algorithm proposed in \cite{stretcu2015multiple}, with code available online, which is an important part of that 
framework. It has lower accuracy than the full system in~\cite{stretcu2015multiple}, but it is much faster, running at $50-100$ fps. At this speed we can produce one million unsupervised soft segmentations in a reasonable time of about 5-6 hours.

VideoPCA models the background in video frames with Principal Component Analysis. It finds initial foreground regions as parts of the frames that are not reconstructed well
with the PCA model. Foreground objects are smaller than the background and
have more complex movements, which make the foreground less likely to be captured well
by the first principal components. The initial soft masks  are used to learn color models of foreground and background, which are then improved by independent pixel-wise classification based on color. For more details the reader is invited to consult~\cite{stretcu2015multiple}.

\begin{figure}[t]
\begin{center}
   \includegraphics[width=0.8\linewidth]{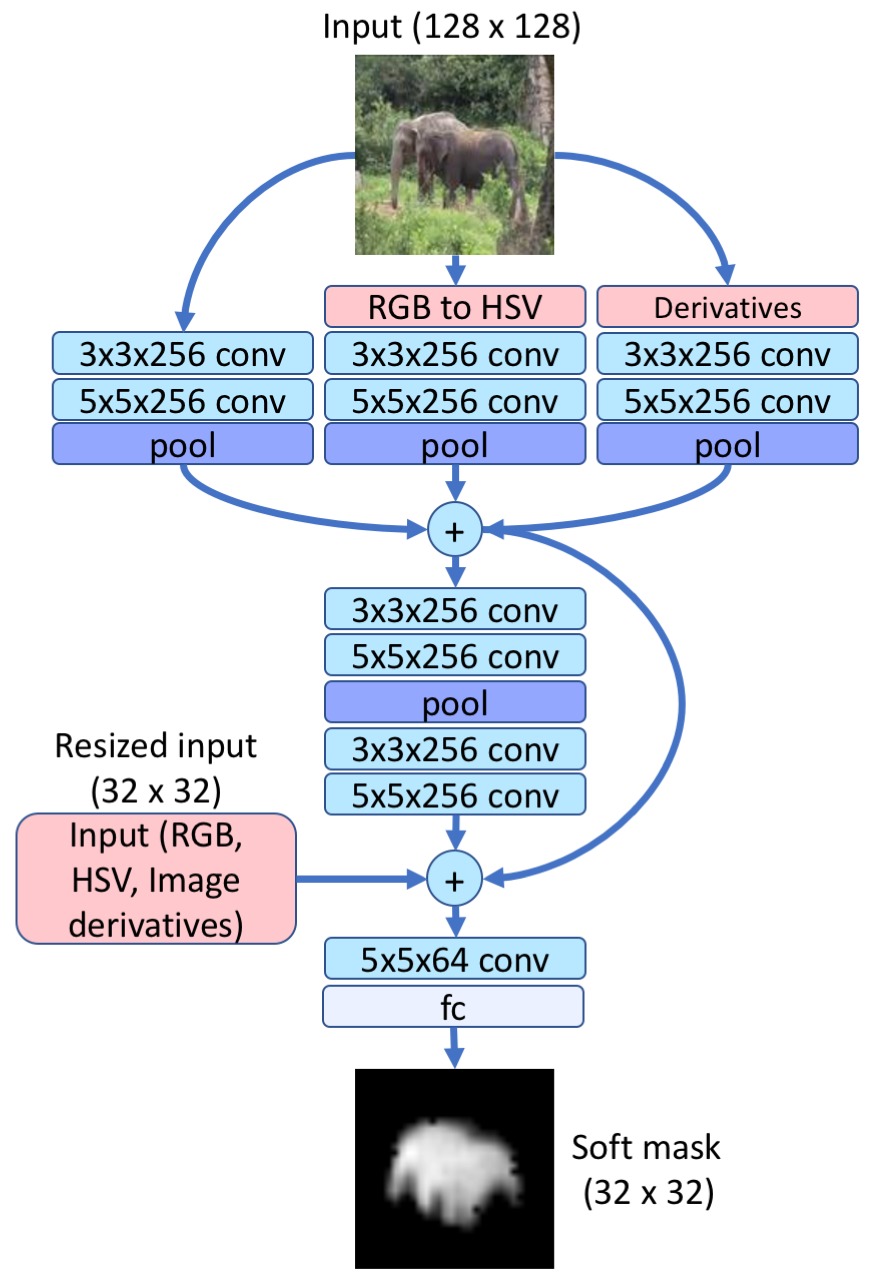}
\end{center}
   \caption{The "student" deep convolutional network architecture that processes single images. It is trained to predict the unsupervised labels given by the teacher pathway, frame by frame. We observed that by adding at the last level the original input and mid-level features (skip connections) and resizing them appropriately, the performance increases.
   }
\label{fig:network}
\end{figure}

\subsection{Student path: single-image segmentation}
\label{sec:cnn}

The student processing pathway (Figure \ref{fig:system}) consists
of a deep convolutional network, with ten layers (seven convolutional, two pooling and one fully connected layer) and skip connections as shown in Figure \ref{fig:network}. 
Skip connections have proved to provide a boost in the network's performance~\cite{raiko2012deep, pinheiro2016learning}. We also observed a slight improvement in our case ($\approx \%1$). The net takes as input a $128\times128$ color image (along with its hue, saturation, derivatives w.r.t. x and y)
and produces a $32\times32$ soft segmentation of the main objects present in the image. 
While it does not identify the particular object classes, it learns from the unsupervised soft-masks provided by the teacher to detect and softly segment the main foreground objects present, regardless of their particular category, one frame at a time. Thus, as shown in experiments, it is also able to detect and segment classes it has never seen before.

We treat foreground object segmentation as a regression problem,
where the soft mask given by the unsupervised video segmentation system acts as the
desired output. Let $\mathbf{I}$ be the input RGB image (a video frame) and $\mathbf{Y}$ be the
corresponding 0-255 valued soft segmentation given by the unsupervised teacher pathway for that particular frame. 
The goal of our network is to predict a soft segmentation mask
$\mathbf{\hat{Y}}$ of width $W=32$ and height $H=32$, that approximates as good as possible the mask 
$\mathbf{Y}$. In other words, for each pixel in the output image, 
we predict a 0-255 value, so that the total difference between 
$\mathbf{Y}$ and $\mathbf{\hat{Y}}$ is minimized. So, given a set of 
$N$ training examples, let $\mathbf{I}^{(n)}$ be the input image (a video frame), ${\mathbf{\hat{Y}}}^{(n)}$ be the 
predicted output mask for $\mathbf{I}^{(n)}$, 
$\mathbf{Y}^{(n)}$ the soft segmentation mask (corresponding to $\mathbf{I}^{(n)}$) and  $\mathbf{w}$ the network parameters.  $\mathbf{Y}^{(n)}$ is produced by the video discoverer by processing the whole video that $\mathbf{I}^{(n)}$ belongs to. Then, our loss is:

\begin{equation}
\label{eq:learning}
L(\mathbf{w})=\frac{1}{N}\sum\limits_{n=1}^N \sum\limits_{p=1}^{W\times H}{(\mathbf{Y}_{p}^{(n)} - \mathbf{\hat{Y}}_{p}^{(n)}(\mathbf{w}, \mathbf{I}^{(n)}))}^2
\end{equation}

where $\mathbf{Y}_{p}^{(n)}$ and $\mathbf{\hat{Y}}_{p}^{(n)}$ denotes the $p$-th pixel from $\mathbf{Y}^{(n)}$, respectively $\mathbf{\hat{Y}}^{(n)}$. We observed that in our tests, the L2 loss performed better than the cross-entropy loss.

We train our network using the Tensorflow~\cite{abaditensorflow} framework 
with the Adam optimizer~\cite{kingma2014adam}. All our models are trained 
end-to-end using a fixed learning rate of 0.001. The training time for a
given model is about 3 days on a Nvidia GeForce GTX 1080 GPU.

\paragraph{Post-processing:}As we stated before, our CNN outputs a $32\times32$ soft mask. In order to be able to fairly compare ourselves against other methods, we have two different post processing steps: 1) bounding box fitting and 2) segmentation refinement. For fitting a box around our soft mask, we first up-sample the $32\times32$ output mask to the original size of the image, then threshold the output, determine the connected components, filter out the small ones and finally fit a tight box around each of the remaining components. 
However, if we are interested in obtaining a fine object segmentation, we use the OpenCV implementation of the GrabCut~\cite{rother2004grabcut} method to refine our soft mask, up-sampled to the original size.

\subsection{Unsupervised soft masks selection}\label{sec:data_selection}

\begin{figure}[t]
\begin{center}
   \includegraphics[width=0.9\linewidth]{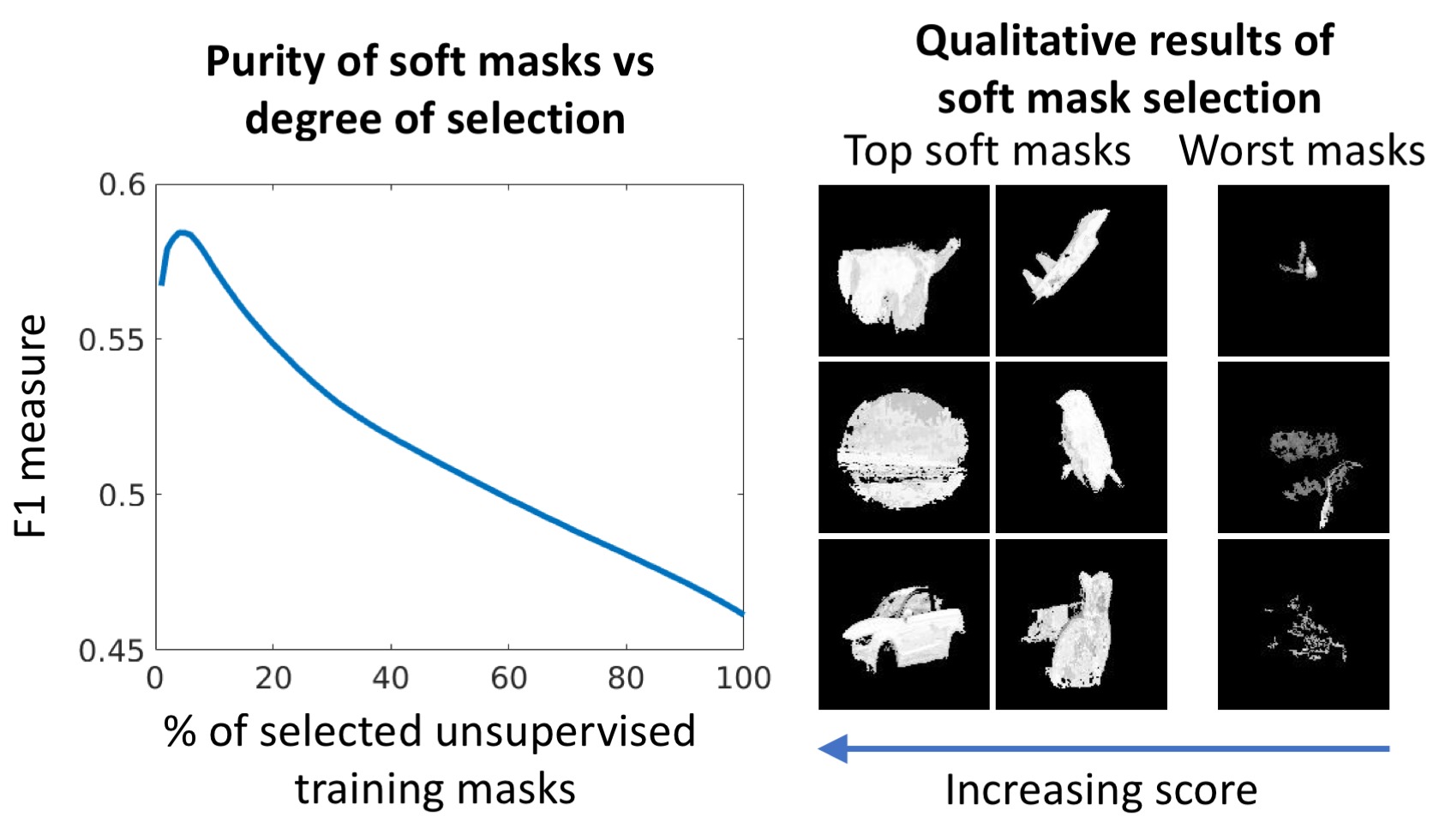}
\end{center}
   \caption{Purity of soft masks vs degree of selection. The more selective we are, the purity of the training frames as compared to the ground truth bounding boxes improves. Note that our selection method is not perfect and some low quality segmentations have high scores or we remove some good segmentations.}
\label{fig:data_augmentation}
\end{figure}

The performance of the single-image processing pathway is influenced by the quality of the 
soft masks provided as labels by the video discovery path. The cleaner and sharper masks provided by the teacher, the more chances the student has to actually learn to segment well the objects in images. VideoPCA used by the video processing path 
usually has good results if the object present in the video stands out against the background scene, in terms of motion and appearance.
However, if the object
is occluded at some point in the video, or if it does not move w.r.t the scene or if it has a very similar appearance to its background, the resulting soft masks might 
be poor.  We used a simple measure of masks quality based on the following observation: when masks are close to the ground truth, the mean of their nonzero values is usually high - which means that when the discoverer is generally confident about a certain mask it is more likely to be closer to the true segmentation.
The mean value of non-zero pixels in the soft mask 
is then used as a score indicator for each segmented frame. 

Next we sort all soft masks in the entire training dataset (e.g. VID~\cite{russakovsky2015imagenet}, YTO~\cite{prest2012learning}) in descending order of their mean score and keep only the top $k$ percent. In this way, we obtain a very simple but completely unsupervised selection method.
In Figure~\ref{fig:data_augmentation} we present the dependency of segmentation performance w.r.t ground truth object boxes (used only for evaluation) vs. the percentile $k$ of masks kept after the automatic selection. In other words, the fewer frames we select the more likely it is that they are correctly segmented. This procedure is not perfect, however, so we sometimes remove good segmentations during this masks selection step.
Even though we can expect to improve the quality of the unsupervised masks by drastically pruning them, the fewer we are left with, the less training data we have, which increases the chance of overfitting. Therefore there is a price to pay. We make up for the losses in training data by augmenting the set of training masks (explained in Section~\ref{sec:data_augmentation}) and by bringing in unlabeled videos from other datasets. Thus, the more selective we are about what masks to accept for training, the more videos we need to collect and process with the teacher pathway, in order to improve generalization.

\subsection{Data augmentation}\label{sec:data_augmentation}

Another drawback of VideoPCA is that it can only detect the main object if it is close to the center of the image. The assumption that the foreground is close to the center is often true and indeed helps that method to produce soft masks with a relatively high precision, but it fails when the object is not in the center, therefore its recall is relatively low.
Our data augmentation procedure also addresses this limitation. This module can be concisely described as follows: scale the input image and the corresponding soft mask given by the video discovery framework at a higher resolution ($160\times160$) and randomly crop $128\times128$ patches from the scaled version. Finally, we down-scale each soft mask to $32\times32$. This would produce slightly larger objects at locations that cover the whole image area, not just the center. As our experiments show, the student net is able to see objects at different locations in the image, unlike its raw teacher, which is strongly biased towards the image center.
Data selection, along with data augmentation of the training set significantly improves unsupervised learning, as shown in the experiments section (Section ~\ref{sec:experiments}).

\begin{figure*}
\begin{center}
   \includegraphics[width=1.0\linewidth]{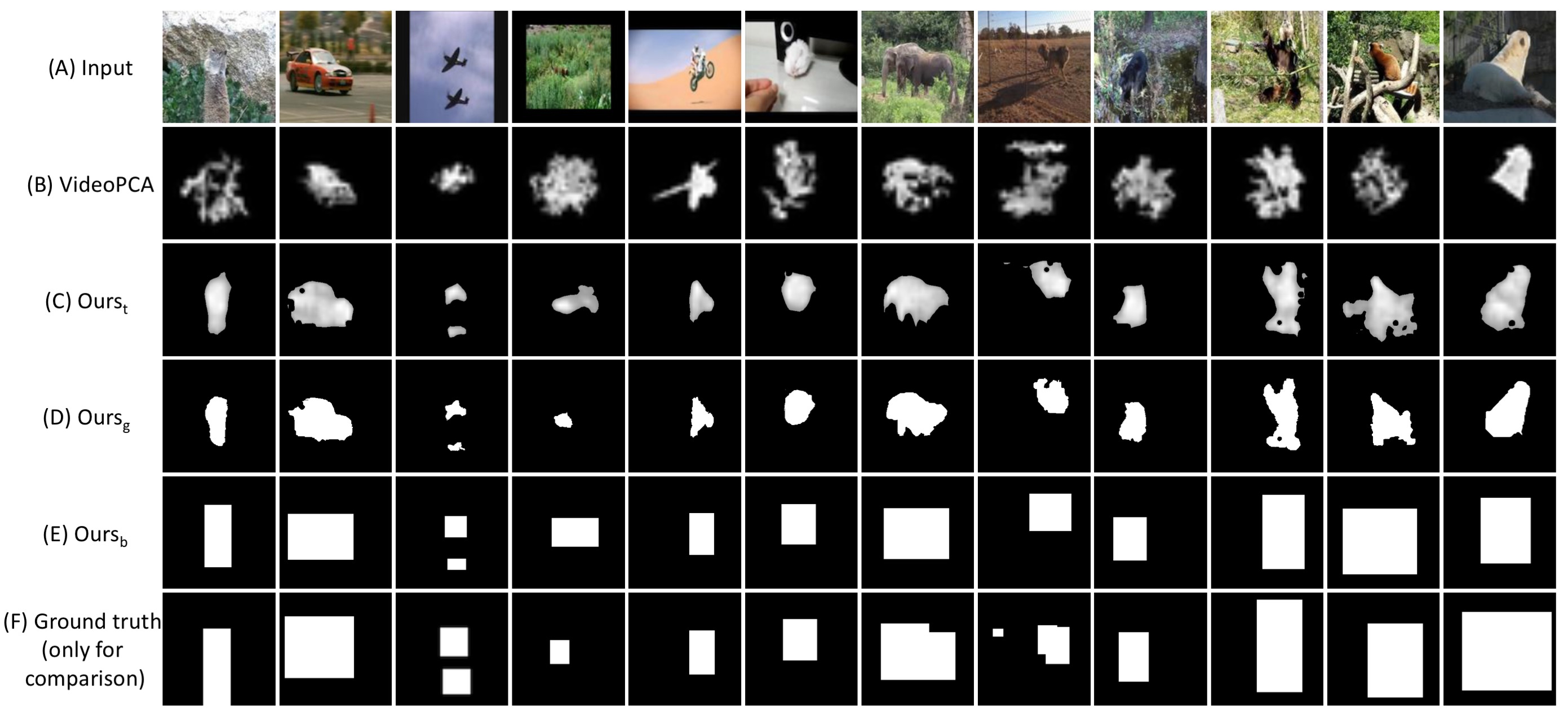}
\end{center}
   \caption{Visual results on the VID dataset~\cite{russakovsky2015imagenet} compared to the teacher method.
A: current frame, B: soft mask produced by VideoPCA~\cite{stretcu2015multiple} for the current frame, after processing the entire video, C: thresholded soft mask produced by our network, D: segmentation mask produced after refining the soft output of our network with GrabCut~\cite{rother2004grabcut}, E: bounding box obtained from the soft segmentation mask; F: ground truth bounding box.}
\label{fig:vid_visual}
\end{figure*}

\section{Experimental analysis}
\label{sec:experiments}

The experiments we conducted aim to highlight various aspects about the performance of our method. Firstly, we compare the quality of the segmentations obtained by the feed-forward CNN against its teacher, VideoPCA (Section~\ref{sec:VID}).
Secondly, we tested that adding extra unlabeled videos improves performance (Section~\ref{sec:more_data}). Finally, we compare the performance of our unsupervised system to state of the art approaches in the field for object discovery in video, testing on the YouTube Objects Dataset~\cite{prest2012learning} benchmark and object discovery in images, testing on the Object Discovery in Internet images~\cite{rubinstein2013unsupervised} benchmark.
(Section~\ref{sec:comparisons}). 

\subsection{Unsupervised learning from ImageNet}
\label{sec:VID}
It is a well known fact that the performance of a convolutional network strongly depends on the amount of data used for training. Because of this, we chose to use as our
primary training dataset the ImageNet Object Detection from Video (VID) dataset~\cite{russakovsky2015imagenet}. VID is one of the largest video datasets publicly available, being fully annotated with ground truth bounding boxes. The large set of annotations available allowed us to have a thorough evaluation of our unsupervised system. The dataset consists of about 4000 videos, having a total of
about 1.2M frames. The videos contain objects that
belong to 30 different classes. Each frame could have zero, one or multiple objects annotated. 
The benchmark challenge associated with this dataset
focuses on the supervised object detection and recognition problem,
which is different from the problem that we tackle here. Our system is not trained 
to identify different object categories. On the VID dataset we evaluated the student CNN against
its teacher pathway. We measure performance of soft-masks by maximum F-measure computed w.r.t ground truth bounding box, by considering pixels inside the bounding box as true positives and those outside as true negatives. This simple metric allows us to use the soft masks directly, without any post-processing steps.

\begin{table}[t]
\begin{center}
\begin{tabular}{|l|c|c|}
\hline
Method & F1 measure & Dataset \\
\hline\hline
VideoPCA~\cite{stretcu2015multiple} & 41.83 & - \\
\hline
Baseline & 51.17 & VID \\
Baseline & 51.9 & VID + YTO \\
Refined & 52.51 & VID \\
Data selection 5\% & 53.20 & VID\\
Data selection 10\% & 53.82 & VID\\
Data selection 30\% & 53.67 & VID\\
Data selection 10\% & 54.53 & VID + YTO\\
\hline
\end{tabular}
\end{center}
\caption{Results on the VID dataset~\cite{russakovsky2015imagenet}. The "dataset" column refers to the datasets used for training the student network. Our baseline model is represented by a classic CNN having only the RGB image as input and no skip-connections. The refined model is our final student CNN model as presented in Figure \ref{fig:network}.
The data selection entries refer to the percentage of kept soft masks after applying our selection method. All soft masks selection experiments were conducted using the refined model. We want to highlight that the overall system performance improves with the amount of selectivity. This shows that a simple quality measure used for soft mask selection can improve the performance of the CNN image-based pathway and that the data augmentation module makes up for the frames lost during the selection process.}
\label{tab:vid}
\end{table}

We tested our unsupervised system on the validation split of the VID dataset. As it can be seen from Table~\ref{tab:vid} our system outperforms its teacher (VideoPCA) by a very significant margin.
Also, in Figure~\ref{fig:vid_visual} we present some qualitative results on this dataset as compared to VideoPCA. We can see that the masks produced by VideoPCA are of lower quality, having holes, non-smooth boundaries and strange shapes that are far from the idea of "objectness"~\cite{alexe2010object}. In contrast, the student learns general shape and appearance characteristics of objects in images reminding of the grouping principles governing the basis of visual perception as studied by the Gestalt psychologists~\cite{rock1990gestalt}. Note that object masks produced by the student are simpler, with very few holes, have nicer and smoother shapes 
and capture well the figure-ground contrast and organization. Another interesting observation is that the network is able to detect multiple objects, a feature that is less commonly achieved by the teacher.
 
\subsection{Adding more data}
\label{sec:more_data}

We also tested how adding more unlabeled data 
affects the overall performance of our system. 
Therefore, we added 
the Youtube Objects(YTO) dataset
to the existing VID dataset. 
The YTO dataset
is a weakly annotated dataset that consists of about 2500 videos, having
a total of about 720K frames, divided into 10 classes.
Adding more unlabeled videos (from YouTube Objects dataset, without annotations) to the unsupervised training set clearly improves performance as reported in Tables \ref{tab:object_discovery_pj},
\ref{tab:vid} and \ref{tab:yto}. The capacity of our system to improve its performance in the presence of unlabeled data, without degradation or catastrophic forgetting is mainly due to the robustness of the teacher pathway combined with data selection and augmentation, in conjunction with the tendency of the single-image CNN net to improve over its teacher.

As it comes to the soft mask selection, our experiments show that we obtain the best overall results by using the top 10\% soft masks with data augmentation. Because of this, all other experiments are conducted using this setup for each dataset.

\subsection{Comparisons with other methods}
\label{sec:comparisons}

\begin{figure*}
\begin{center}
   \includegraphics[width=1.0\linewidth]{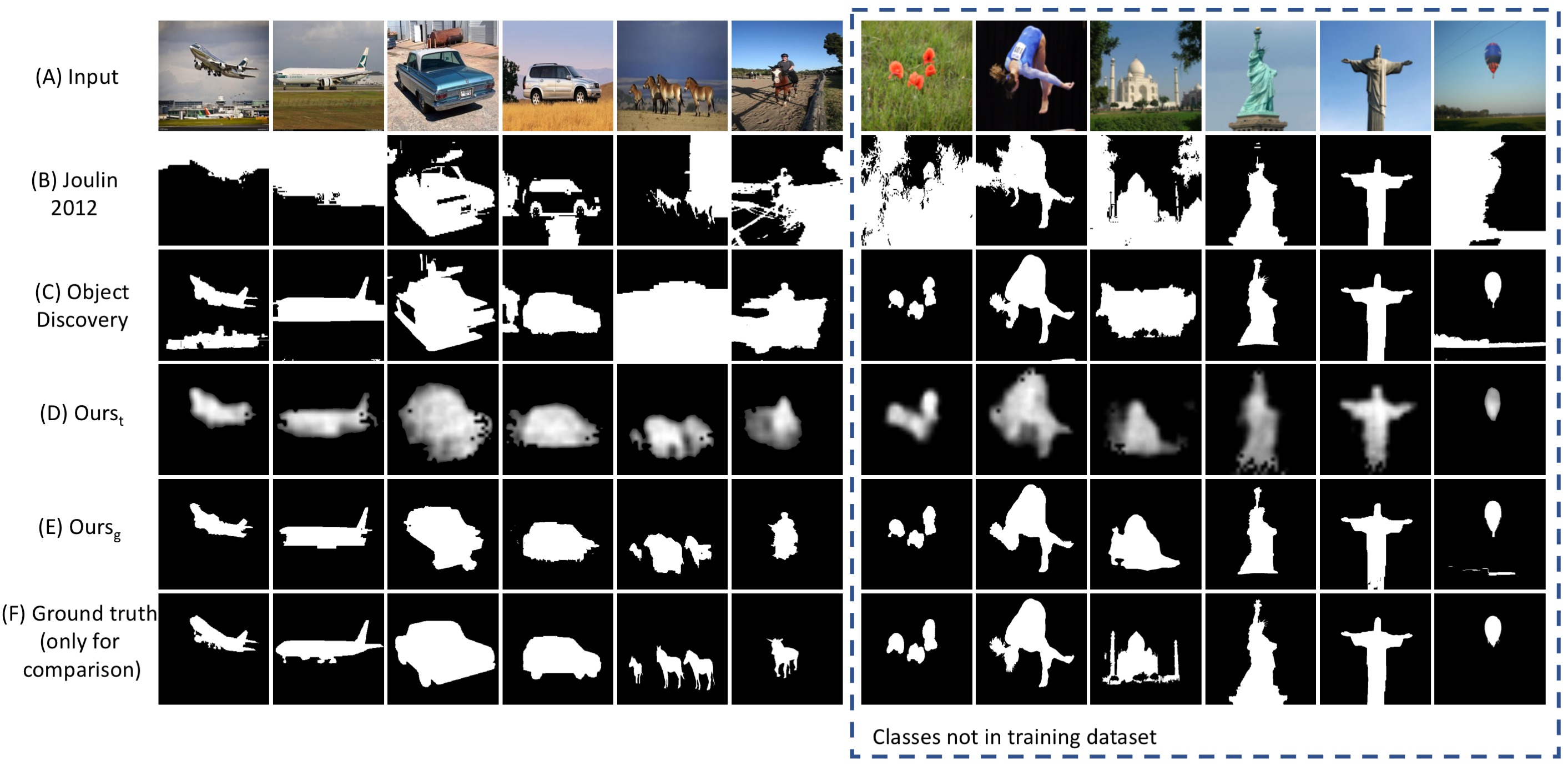}
\end{center}
   \caption{Visual results on the Object Discovery dataset. A: input image, B: segmentation obtained by~\cite{joulin2012multi}, C: segmentation obtained by~\cite{rubinstein2013unsupervised}, D: thresholded soft mask produced by our network, E: segmentation mask produced after refining the soft output of our network with GrabCut~\cite{rother2004grabcut}, F: ground truth segmentation. }
\label{fig:internet_visual}
\end{figure*}

\paragraph{Single image discovery methods}
Next, we compare our unsupervised system with state of the art methods designed for the task of object discovery in collections of images, that might contain one or a few main object categories of interest. A representative current benchmark in this sense is the Object Discovery in Internet images dataset. This set contains Internet images and it is annotated with high detail segmentation masks. In order to enable comparison with previous methods, we use the 100 images subsets.

The methods evaluated on this dataset, in the literature, aim to either discover the bounding box of the main object in given image, or its fine segmentation mask. We evaluate our system on both. Different from other methods, we do not need a collection of images during testing, since each image is processed independently by our system, at test time. Therefore, our performance is not affected by the structure of the image collection or the number of classes of interest being present in the collection.

\begin{table}[t]
\begin{center}
\begin{tabular}{|l|c|c|c|c|}
\hline
Method & Airplane & Car & Horse & Avg\\
\hline\hline
~\cite{kim2011distributed} & 21.95 & 0.00 & 16.13 & 12.69\\
~\cite{joulin2010discriminative} & 32.93 & 66.29 & 54.84 & 51.35 \\
~\cite{joulin2012multi} & 57.32 & 64.04 & 52.69 & 58.02\\
~\cite{rubinstein2013unsupervised} & 74.39 & 87.64 & 63.44 & 75.16\\
~\cite{tang2014co} & 71.95 & 93.26 & 64.52 & 76.58\\
~\cite{cho2015unsupervised} & 82.93 & 94.38 & \color{red}\textit{75.27} & 84.19\\
~\cite{cho2015unsupervised}(mixed-class) & 81.71 & 94.38 & 70.97 & 82.35\\
\hline

Ours\textsubscript{VID} & \textbf{93.90} & 93.26 & 70.97 &  \textbf{86.04}\\
Ours\textsubscript{VID+YTO} & 87.80 & \textbf{95.51} & \textbf{74.19} & 85.83 \\
\hline
\end{tabular}
\end{center}
\caption{Results on the Object Discovery in Internet images~\cite{rubinstein2013unsupervised} dataset (CorLoc metric). Ours\textsubscript{VID} represents our network trained using the VID dataset (with 10\% selection), while Ours\textsubscript{VID+YTO} represents our network trained on VID and YTO datasets (with 10\% selection).}
\label{tab:object_discovery}
\end{table}

\begin{table}[t]
\begin{center}
\begin{tabular}{|c|c|c|c|c|c|c|}
\hline
\multirow{2}{*}{} & \multicolumn{2}{c|}{Airplane} & \multicolumn{2}{c|}{Car} & \multicolumn{2}{c|}{Horse} \\
\cline{2-7}
& P & J & P & J & P & J \\
\hline\hline
\cite{kim2011distributed} & 80.20 & 7.90 & 68.85 & 0.04 & 75.12 & 6.43 \\
\cite{joulin2010discriminative} & 49.25 & 15.36 & 58.70 & 37.15 & 63.84 & 30.16 \\
\cite{joulin2012multi} & 47.48 & 11.72 & 59.20 & 35.15 & 64.22 & 29.53 \\
\cite{rubinstein2013unsupervised} & 88.04 & 55.81 & 85.38 & 64.42 & 82.81 & 51.65 \\
\cite{chen2014enriching} & 90.25 & 40.33 & \textbf{87.65} & 64.86 & 86.16 & 33.39 \\

\hline
Ours\textsubscript{1} & 90.92 & \textbf{62.76} & 85.15 & 66.39 & \textbf{87.11} & 54.59 \\

Ours\textsubscript{2} & \textbf{91.41} & 61.37 & 86.59 & \textbf{70.52} & 87.07 & \textbf{55.09} \\

\hline
\end{tabular}
\end{center}
\caption{Results on the Object Discovery in Internet images~\cite{rubinstein2013unsupervised} dataset (P, J metric). Ours\textsubscript{1} represents our network trained using the VID dataset (with 10\% selection), while Ours\textsubscript{2} represents our network trained on VID and YTO datasets (with 10\% selection). We observe that Ours\textsubscript{2} has better results with mean P of \textbf{88.36} and mean J of \textbf{62.33} compared to Ours\textsubscript{1} (mean P: 87.73, mean J: 61.25).}
\label{tab:object_discovery_pj}
\end{table}

For evaluating the detection of bounding boxes 
the most used metric is CorLoc defined
as the percentage of images correctly localized according to the PASCAL criterion:$\frac{B_p \cap B_{GT}}{B_p \cup B_{GT}} \geq 0.5$,
where $B_P$ is the predicted bounding box and $B_{GT}$ is the ground truth bounding box. In
Table~\ref{tab:object_discovery} we present the performance of our method as compared to other
unsupervised object discovery methods in terms of CorLoc on the Object Discovery dataset. We compare our predicted box against the tight box fitted around the ground-truth segmentation as done in \cite{cho2015unsupervised,tang2014co}.
Our system can be considered in the mixed class category: it does not depend on the structure of the image collection. It treats each image independently. The performance of the other algorithms degrades as the number of main categories increases in the collection (some are not even tested by their authors on the mixed-class case).

\begin{figure*}
\begin{center}
   \includegraphics[width=1.0\linewidth]{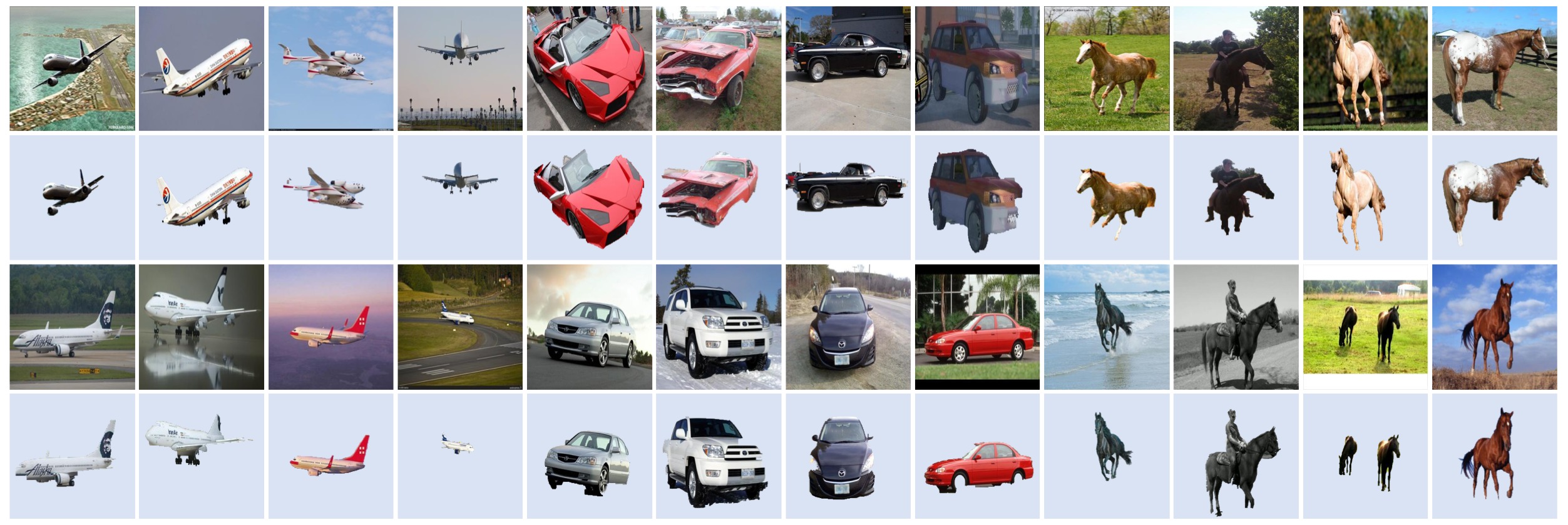}
\end{center}
   \caption{Qualitative results on the Object Discovery in Internet images~\cite{rubinstein2013unsupervised} dataset. For each example we show the input RGB image (first and third row) and immediately below (second and fourth row) our refined segmentation result obtained by applying GrabCut on the soft segmentation mask predicted by our network. Note that our method produces good quality segmentation results, even in cases with cluttered background.}
\label{fig:visual}
\end{figure*}

\begin{table*}
\begin{center}
\begin{tabular}{|l|*{10}{c|}|c|c|c|}
\hline
Method & Aero & Bird & Boat & Car & Cat & Cow & Dog & Horse & Mbike & Train & Avg & Time & Version \\
\hline\hline
~\cite{prest2012learning} & 51.7 & 17.5 & 34.4 & 34.7 & 22.3 & 17.9 & 13.5 & 26.7 & 41.2 & 25.0 & 28.5 & N/A & \multirow{4}{*}{v1~\cite{prest2012learning}} \\
~\cite{papazoglou2013fast} & 65.4 & 67.3 & 38.9 & 65.2 & 46.3 & 40.2 & 65.3 & 48.4 & 39.0 & 25.0 & 50.1 & 4s & \\
~\cite{jun2016pod} & 64.3 & 63.2 & 73.3 & \textbf{68.9} & 44.4 & 62.5 & 71.4 & 52.3 & \textbf{78.6} & 23.1 & 60.2 & N/A & \\
\cline{1-13}
Ours\textsubscript{VID} & 69.8 & 59.7 & 65.4 & 57.0 & 50.0 & \textbf{71.7} & \textbf{73.3} & 46.7 & 32.4 & 34.9 & 56.1 & 0.04s & \\
Ours\textsubscript{VID+YTO} & \textbf{77.0} & \textbf{67.5} & \textbf{77.2} & 68.4 & \textbf{54.5} & 68.3 & 72.0 & \textbf{56.7} & 44.1 & \textbf{34.9} & \textbf{61.6} & 0.04s & \\
\hline
\hline
Ours\textsubscript{VID+YTO} & 75.7 & 56.0 & 52.7 & 57.3 & 46.9 & 57.0 & 48.9 & 44.0 & 27.2 & 56.2 & 52.2 & 0.04s & v2.2~\cite{kalogeiton2016analysing} \\
\hline
\end{tabular}
\end{center}
\caption{Results on Youtube Objects dataset~\cite{prest2012learning}. Ours\textsubscript{VID} represents our network trained using the VID dataset (with 10\% selection), while Ours\textsubscript{VID+YTO} represents our network trained on VID and YTO datasets (with 10\% selection). Note that our system has a significantly lower test time than~\cite{papazoglou2013fast} which we estimate that is the fastest method.}
\label{tab:yto}
\end{table*}

\begin{table*}[ht]
\begin{center}
\begin{tabular}{|l|*{10}{c|}|c|}
\hline
Method & Aero & Bird & Boat & Car & Cat & Cow & Dog & Horse & Mbike & Train & Avg \\
\hline\hline
Whole method~\cite{stretcu2015multiple} & 38.3 & \textbf{62.5} & 51.1 & 54.9 & \textbf{64.3} & 52.9 & 44.3 & 43.8 & \textbf{41.9} & \textbf{45.8} & 49.9 \\
Teacher(VideoPCA)~\cite{stretcu2015multiple} + tight box & 69.6 & 55.8 & 64.9 & 50.5 & 44.8 & 43.2 & 48.6 & 37.4 & 17.9 & 22.0 & 45.5 \\
\hline
Ours & \textbf{71.9} & 61.5 &\textbf{75.4} & \textbf{70.3} & 53.0 & \textbf{59.3} & \textbf{70.6} & \textbf{56.0} & 37.9 & 39.0 & \textbf{59.5} \\
\hline
\end{tabular}
\end{center}
\caption{Results on the whole (training + testing) YouTube Objects dataset~\cite{prest2012learning}. Our system outperforms both VideoPCA (used in the teacher pathway) and the full method from~\cite{stretcu2015multiple} by very significant margins (about $10\%$ and $14\%$, respectively).}
\label{tab:yto_all}
\end{table*}

\begin{figure*}
\begin{center}
   \includegraphics[width=1.0\linewidth]{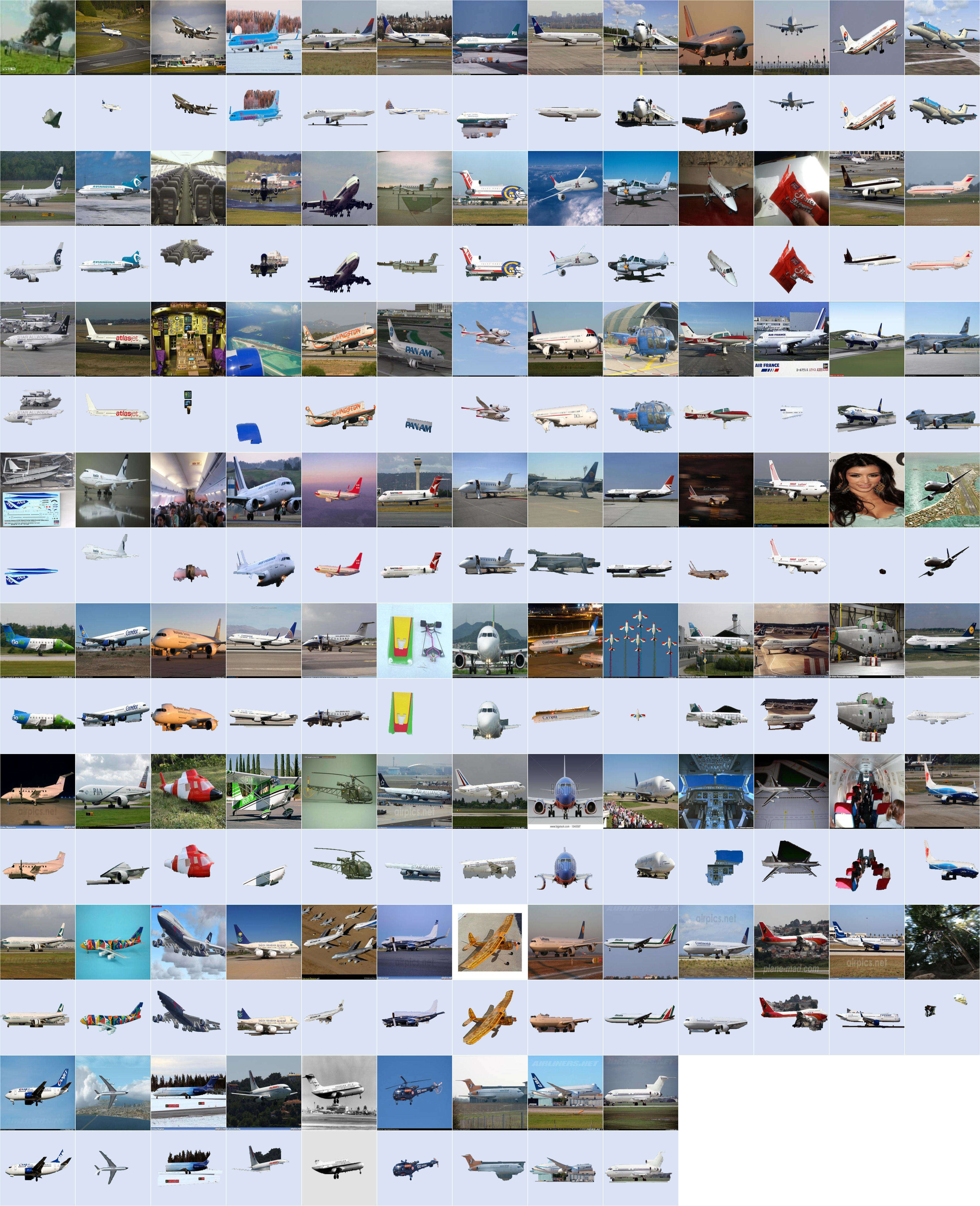}
\end{center}
   \caption{Qualitative results on the 100 airplane image subset from Object Discovery in Internet images~\cite{rubinstein2013unsupervised} dataset.}
\label{fig:all_airplane}
\end{figure*}

\begin{figure*}
\begin{center}
   \includegraphics[width=1.0\linewidth]{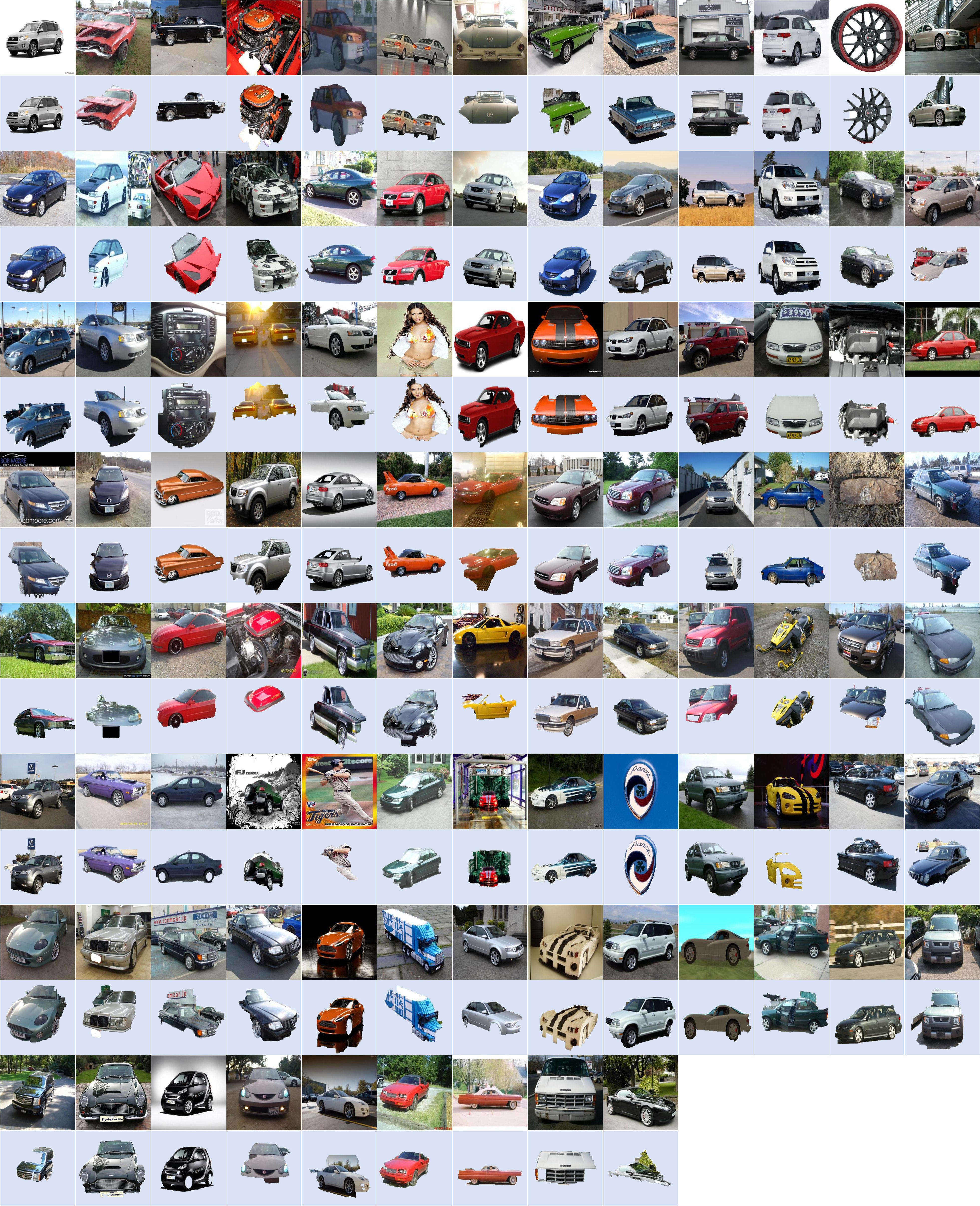}
\end{center}
   \caption{Qualitative results on the 100 car image subset from Object Discovery in Internet images~\cite{rubinstein2013unsupervised} dataset.}
\label{fig:all_car}
\end{figure*}

\begin{figure*}
\begin{center}
   \includegraphics[width=1.0\linewidth]{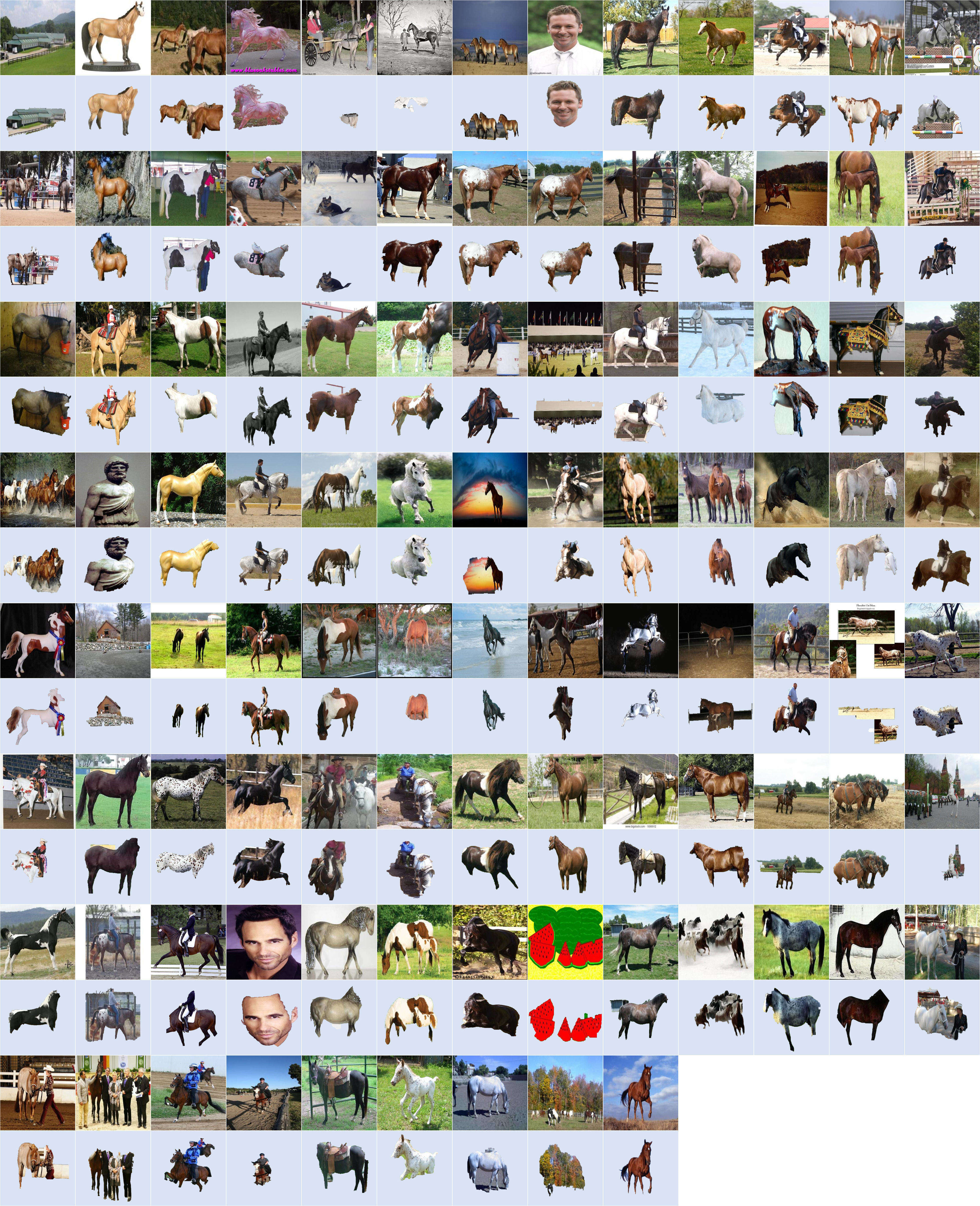}
\end{center}
   \caption{Qualitative results on the 100 horse image subset from Object Discovery in Internet images~\cite{rubinstein2013unsupervised} dataset.}
\label{fig:all_horse}
\end{figure*}

We obtain state of the art
results on all classes (in the mixed class case), improving by a significant margin over the method of \cite{cho2015unsupervised} in the mixed class case. When the method in \cite{cho2015unsupervised} is allowed to see a collection of images that are limited to a single majority class, its performance improves and outperforms ours on one class. However, the comparison is not truly appropriate since our method has no other information necessary besides the input image, at test time.

We also tested our system on the task of fine foreground object segmentation and compared to the best performers in the literature on the Object Discovery dataset in Table~\ref{tab:object_discovery_pj}. For refining our soft masks we apply the GrabCut method, as it is available in OpenCV. We evaluate based on the same
P, J evaluation metric as described by Rubinstein \etal ~\cite{rubinstein2013unsupervised} - the higher P and J, the better. P refers to the per pixel precision, while J is the Jaccard similarity (the intersection over
union of the result and ground truth segmentations). In Figure~\ref{fig:visual} and~\ref{fig:internet_visual} we present some qualitative samples from each class, while in Figure~\ref{fig:all_airplane}, ~\ref{fig:all_car}, ~\ref{fig:all_horse} we present the qualitative results on all the images that we tested.

\paragraph{Video discovery methods} 

We also performed comparisons with methods specifically designed for object discovery in video. For this, we choose the YouTube Objects Dataset and compared to the best performers on this dataset in the literature (Table \ref{tab:yto}). Evaluations are conducted on both 
versions of YouTube Objects dataset,
YTOv1~\cite{prest2012learning} and YTOv2.2~\cite{kalogeiton2016analysing}. On YTOv1 we follow the same experimental setup as~\cite{jun2016pod, prest2012learning}, by running experiments only on the training videos. It is important to stress out again, the fact that, while the methods presented here for comparison have access to whole video shots, ours only needs a single image at test time. Despite this limitation, our method outperforms the others on 8 out of 10 classes and has the best overall average performance. It is also important to note that our CNN feed-forward net processes each image in 0.04 sec, being at least one to two orders of magnitude faster than all other methods (see Table ~\ref{tab:yto}). It is also important to note that in all our comparisons, while our system is faster at test time, it takes much longer during its unsupervised training phase and requires large quantities of unsupervised training data.

In Table~\ref{tab:yto_all} we report additional experiments, on all annotated frames from YouTube Objects in order to compare with the full system of~\cite{stretcu2015multiple}, where VideoPCA was introduced. We also report comparisons with VideoPCA alone on the same train+test split. For VideoPCA we also fitted a tight bounding box. For~\cite{stretcu2015multiple} we report the results presented in their paper.

\section{Conclusions and Future Work}\label{sec:conclusions}

We have shown in extensive experiments that it is possible to use a relatively simple method
for unsupervised object discovery in video to train a powerful deep neural network for detection and segmentation of objects in single images. The result is interesting and encouraging and shows how
a system could learn in a completely unsupervised fashion, general visual characteristics that predict well the presence and shape of objects in images. The network essentially discovers appearance object features from single images, at different levels of abstraction, that are strongly correlated with the spatiotemporal consistency of objects in video.

The student network, during the unsupervised training phase is also able to significantly outperform its teacher, by learning such general "objectness" characteristics that are well beyond the capabilities of its teacher. These characteristics include good form, closure, smooth contours, as well as contrast with its background. What the simpler teacher discovers over time, the deep, complex student is able to learn across several layers of image features at different levels of abstraction. Thus, our unsupervised learning model, tested in extensive experiments, brings a valuable contribution to the unsupervised learning problem in vision research.

\paragraph{Acknowledgements:} This work was supported by UEFISCDI, under project PN-III-P4-ID-ERC-2016-0007.

{\small
\bibliographystyle{ieee}
\bibliography{bib_arXiv_CBL}
}

\end{document}